\documentclass[letterpaper]{article} 
\usepackage{aaai2027}  
\usepackage[hyphens]{url}  
\usepackage{graphicx} 
\def\UrlFont{\rm} 
\usepackage{natbib} 
\usepackage{caption} 
\usepackage{algorithm}
\usepackage{algpseudocode}
\usepackage{booktabs}
\usepackage{multirow}
\usepackage{amsmath}
\usepackage{amssymb}
\usepackage{placeins}

\title{GIFT: Geometry-Invariant Fine-Tuning for Non-Lambertian Monocular Depth Estimation}
\author{
    Xianghui Fan\textsuperscript{\rm 1,\rm 2},
    Zhaoyu Chen\textsuperscript{\rm 3},
    Bingqian Wu\textsuperscript{\rm 4},
    Dayu Li\textsuperscript{\rm 4},\\
    Xin Zeng\textsuperscript{\rm 1,\rm 2},
    Cui Huanran\textsuperscript{\rm 1,\rm 2},
    Guangzhen Xu\textsuperscript{\rm 1,\rm 2},
    Xiangru Huang\textsuperscript{\rm 4},
    Hang Yang\textsuperscript{\rm 1,\rm 2}\corresponding
}
\affiliations{
    \textsuperscript{\rm 1}Changchun Institute of Optics, Fine Mechanics and Physics, Chinese Academy of Sciences\\
    \textsuperscript{\rm 2}University of Chinese Academy of Sciences\\
    \textsuperscript{\rm 3}Fudan University\\
    \textsuperscript{\rm 4}Westlake University
}

\begin{document}

\maketitle

\begin{abstract}
Monocular depth foundation models, benefiting from large-scale synthetic training data, have demonstrated strong generalization. However, they often hallucinate depth on non-Lambertian surfaces, estimating reflected content in mirrors or transmitted content behind glass rather than the physical surface itself. Adapting these models with real-world data is challenging because conventional depth sensors are also unreliable in such regions. We observe that while the appearance of a non-Lambertian surface varies with its reflected or transmitted environment, its underlying geometry remains unchanged. Based on this observation, we propose GIFT (Geometry-Invariant Fine-Tuning), a parameter-efficient post-training framework that requires no measured depth labels. We collect groups of RGB images under controlled appearance changes while keeping the camera and target geometry fixed. GIFT exploits geometric invariance across these observations to suppress non-Lambertian depth hallucinations while retaining general depth estimation capability. We further construct a controlled benchmark that evaluates non-Lambertian depth recovery, robustness to appearance changes, and performance retention in other regions. Experiments on our benchmark and an independent real-world dataset demonstrate that GIFT improves depth prediction for mirrors and transparent objects while largely preserving the base model’s performance, providing a practical and low-cost approach for adapting monocular depth foundation models to non-Lambertian scenes.
\end{abstract}

\section{Introduction}

Monocular depth estimation has progressed from dataset-specific predictors to
models that transfer across diverse scenes and, in some cases, recover metric
depth without camera metadata
\citep{ranftl2022midas,yang2024depthanythingv2,bochkovskiy2025depthpro}.
This progress nevertheless relies on regularities between image appearance
and physical geometry. Mirrors and transparent surfaces violate these
regularities: the observed radiance may describe a reflected or transmitted
scene rather than the surface that generated the image. A monocular model may
therefore place a reflected room behind a wall or recover the background
behind a glass pane instead of the pane itself. The same materials also cause
substantial errors in commonly used depth measurements, making both inference
and supervision difficult
\citep{tan2021mirror3d,ramirez2024booster,jiang2024transparentreview}.

Recent methods mainly replace unreliable measured depth with constructed
supervision. Some rely on synthetic or generated non-Lambertian data
\citep{zhang2025nonlambertian,tosi2024diffusionrobustdepth,
	wen2025layereddepth}; DKT further adapts a video diffusion model using
synthetic RGB--depth pairs for transparent-object depth estimation
\citep{xu2025dkt}. Others modify the input appearance: Depth4ToM fills masked
non-Lambertian regions to construct virtual labels, while generative
opacification methods replace transparent objects with opaque counterparts
\citep{costanzino2023depthtom,wang2026seeclear}. However, these approaches may
suffer from synthetic-to-real gaps or geometry distortions introduced by
image generation and editing, and dedicated generation modules can add
training or inference cost.

A simple physical observation motivates our approach: although the appearance
of a non-Lambertian object can vary substantially with its environment, its
underlying geometry remains unchanged. Accordingly, a depth foundation model
should produce stable geometry predictions for the same object under such
appearance variations. We therefore intervene on the reflected or transmitted
content to alter the target appearance while preserving its geometry, and
minimize the discrepancy between the resulting depth predictions. This
cross-appearance disagreement provides a depth-GT-free self-supervised signal
for adapting the model to non-Lambertian surfaces.

Based on this principle, we introduce GIFT (Geometry-Invariant Fine-Tuning),
a fine-tuning framework for non-Lambertian depth estimation, together with
the GIFT-Intervention dataset (GIFT-I) that enables its depth-GT-free
training. GIFT-I contains 353 training groups with 3,268 RGB images and a
disjoint validation set of 41 groups. Within each group, the camera pose and
target geometry remain fixed while the reflected or transmitted content is
varied, and target masks identify the regions in which geometric invariance
should hold.

GIFT combines two complementary objectives during fine-tuning. A
geometry-invariance loss constrains target-region depth variation across all
valid image pairs within an intervention group. Because consistency alone
admits group-shared degenerate solutions, such as predicting a constant depth
map, a frozen-model self-distillation loss further constrains predictions
outside the target and preserves the pretrained depth prior. For efficient
adaptation, we employ LoRA \citep{hu2022lora} together with prediction reuse
and group-wise micro-batching, enabling rapid fine-tuning on a single
consumer-grade GPU. We further introduce geometry-consistency error (GCE), a
depth-GT-free metric that directly measures the invariance of predicted
geometry under controlled appearance changes and partially reflects the
model's non-Lambertian depth estimation performance.

We apply GIFT to three representative depth foundation models: Depth Anything
V2 \citep{yang2024depthanythingv2}, VGGT-1B \citep{wang2025vggt}, and
Metric3D V2 \citep{hu2024metric3dv2}. Experiments on the GIFT-I validation set
and the Booster benchmark demonstrate clear improvements in non-Lambertian
depth estimation.

Our contributions are:
\begin{itemize}
	\item We construct the grouped GIFT-I dataset, enabling the training and
	validation of non-Lambertian depth estimation without measured depth
	labels.
	
	\item We propose GIFT, which combines complete-group geometry-invariance
	supervision with frozen-model self-distillation to improve
	non-Lambertian depth estimation while preventing prediction collapse and
	preserving the pretrained depth prior.
	
	\item We design an efficient LoRA-based group-wise optimization strategy
	with prediction reuse and micro-batching, enabling rapid adaptation on a
	single consumer-grade GPU.
	
	\item We validate GIFT on Depth Anything V2, VGGT-1B, and Metric3D V2,
	demonstrating improvements on GIFT-I and Booster.
\end{itemize}

\section{Related Work}

\paragraph{Monocular depth foundation models.}
MiDaS and DPT established robust cross-dataset relative depth
\citep{ranftl2022midas,ranftl2021dpt}. Depth Anything and its second version
scale this direction with large unlabeled corpora
\citep{yang2024depthanything,yang2024depthanythingv2}, while Metric3D,
UniDepth, and Depth Pro target metric geometry
\citep{yin2023metric3d,piccinelli2024unidepth,bochkovskiy2025depthpro}.
Diffusion-based estimators provide another strong geometric prior
\citep{ke2024marigold,fu2024geowizard}. VGGT jointly predicts cameras, depth,
point maps, and tracks with a feed-forward transformer
\citep{wang2025vggt}. GIFT is not a new backbone, but a post-training
procedure designed to adapt depth foundation models with different geometric
priors.

\paragraph{Non-Lambertian depth completion and estimation.}
Depth completion methods typically recover transparent-object geometry from
RGB-D observations with missing or corrupted sensor depth. ClearGrasp
predicts transparent masks, surface normals, and occlusion boundaries to
refine the raw depth, while A4T combines hierarchical affordance detection
with progressive depth reconstruction for robotic manipulation
\citep{sajjan2020cleargrasp,jiang2022a4t}. TDCNet employs parallel CNN and
Transformer branches to extract complementary features from partial depth
and RGB-D inputs \citep{fan2025tdcnet}. Fan et al.\ further introduce a
self-supervised strategy that simulates transparent-like depth defects on
non-transparent regions and uses the original valid depth as supervision
\citep{fan2025selfsupervised}. ReMake combines instance masks with monocular
depth priors to guide RGB-D depth completion and improve generalization to
real-world grasping scenarios \citep{cheng2026remake}.

Beyond depth completion, Mirror3D refines mirror regions using semantic and
planar cues \citep{tan2021mirror3d}. Booster provides stereo ground truth for
specular and transparent surfaces \citep{ramirez2024booster}, while other
methods exploit contextual cues, layered representations, diffusion priors,
or additional modalities
\citep{liang2023glasswalls,shi2023distillation,
	tosi2024diffusionrobustdepth,wen2025layereddepth}. Depth4ToM constructs
virtual depth supervision by filling masked non-Lambertian regions with
random colors \citep{costanzino2023depthtom}. In contrast, GIFT preserves
the real RGB observations and adapts pretrained monocular depth models by
exploiting the geometry invariance shared within physical intervention
groups, without measured training depth.

\begin{figure*}[t]
	\centering
	\includegraphics[width=\textwidth]{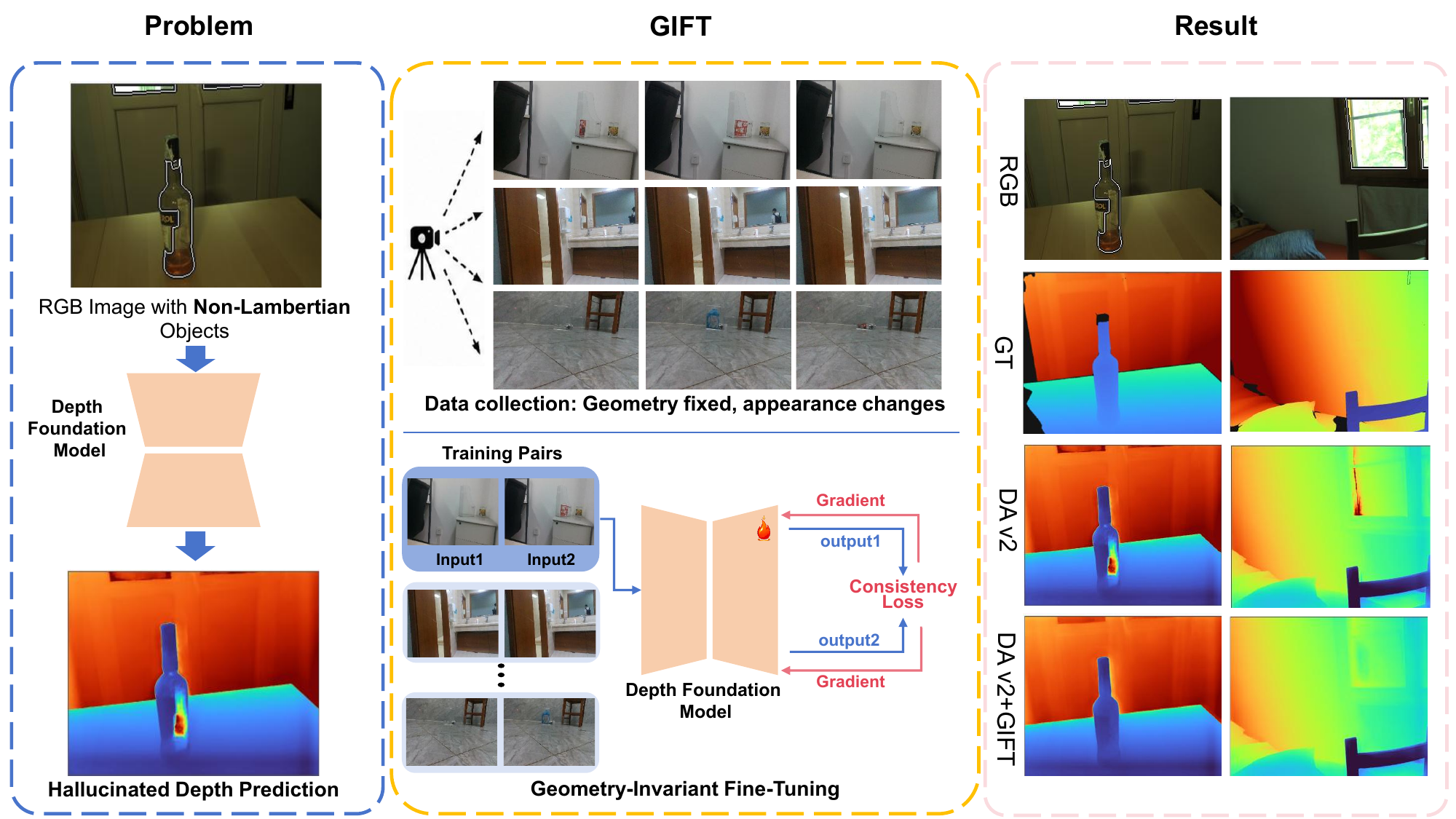}
	\caption{
		Overview of GIFT. A pretrained depth foundation model may interpret
		reflected or transmitted content as physical geometry, producing
		hallucinated depth on non-Lambertian surfaces. GIFT collects grouped
		observations under fixed camera pose and target geometry while varying
		the reflected or transmitted content, and adapts the model by minimizing
		target-region depth disagreement across observations. The adapted model
		suppresses appearance-dependent depth hallucinations and recovers more
		stable surface geometry.
	}
	\label{fig:main}
\end{figure*}

\section{Method}

Figure~\ref{fig:main} summarizes the motivation and overall pipeline of GIFT.
A pretrained depth foundation model may incorrectly recover reflected or
transmitted content instead of the physical non-Lambertian surface. To address
this problem, GIFT first collects intervention groups in which the camera pose
and target geometry remain fixed while the target appearance varies with the
reflected or transmitted environment. It then adapts the pretrained model
using complete-group geometry-invariance supervision, together with
frozen-model self-distillation that preserves the pretrained depth prior
outside the target region. The entire training process requires no measured
depth labels.

\subsection{GIFT-Intervention Dataset}
\label{sec:dataset}

\begin{figure}[t]
	\centering
	\includegraphics[width=\linewidth]{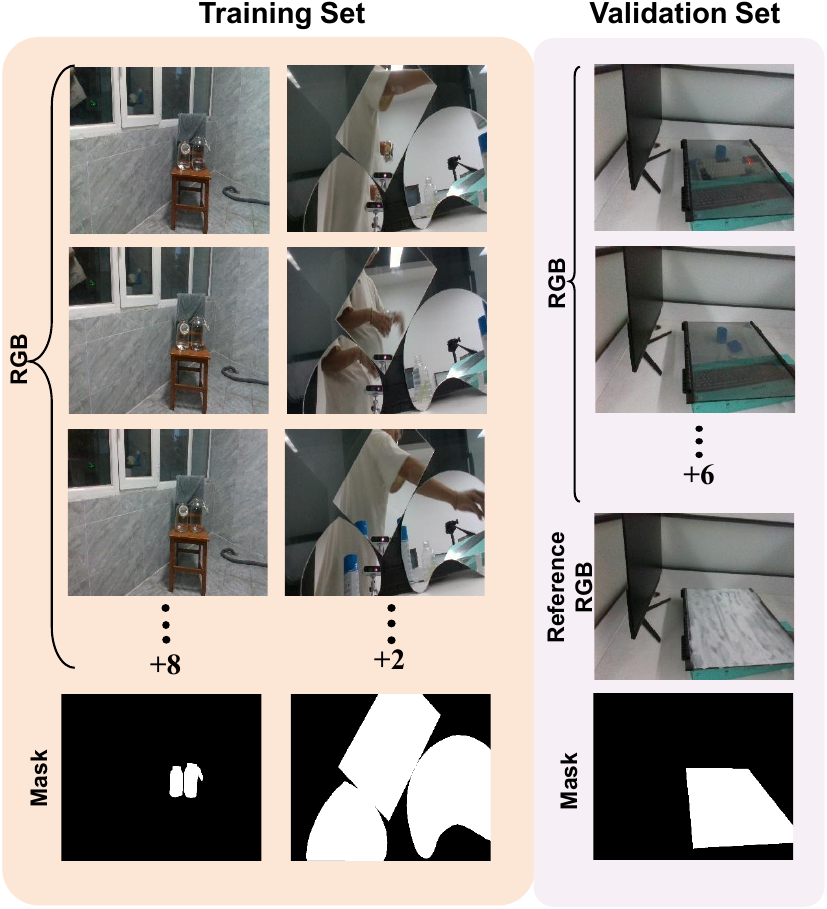}
	\caption{
		Overview of the GIFT-Intervention Dataset. Each training group contains
		RGB images captured under fixed camera pose and target geometry but
		varying non-Lambertian appearance, together with the corresponding
		target-region masks. Each validation group additionally contains a
		surface-treated reference RGB image, obtained using matte spray paint
		or opaque matte stickers, for constructing a proxy depth reference.
		The ``+$N$'' notation indicates that $N$ additional RGB images from the
		same intervention group are not shown.
	}
	\label{fig:gift_dataset}
\end{figure}

To provide supervision without measured depth labels, we construct the
\textbf{GIFT-Intervention Dataset} (\textbf{GIFT-I}), a real-world grouped
RGB dataset for adapting monocular depth models to non-Lambertian surfaces.
As illustrated in Fig.~\ref{fig:gift_dataset}, each image group is captured
using a rigidly mounted camera while the camera pose and target geometry
remain fixed. For transparent objects, we move or replace the content behind
the target; for mirrors, we change the content reflected by the surface.
These interventions produce substantial variations in target appearance
while preserving its physical depth and pixel correspondence. A
target-region mask is manually annotated for each image.

To maintain the fixed-geometry assumption, we apply lightweight image
registration and quality filtering to remove captures with noticeable target
misalignment. Detailed capture settings, intervention procedures for
transparent and mirror targets, registration, filtering, mask annotation,
and proxy-reference construction are provided in the supplementary material,
together with additional GIFT-I groups illustrating the diversity of objects,
materials, and scenes.

The training split contains 353 groups and 3,268 RGB images, with group
sizes ranging from 2 to 55. A disjoint validation split contains 41 groups
with 350 non-Lambertian input images. Each validation group additionally
includes one surface-treated reference RGB image, resulting in 41 reference
images in total.

The reference images are excluded from training and are not used as inputs
to the evaluated model. Instead, predictions produced by the frozen depth
foundation model on these reference images are used only to construct proxy
depth references for assessing target-region recovery. Thus, GIFT-I requires
no measured depth labels for either model adaptation or validation.

\subsection{Geometry-Invariant Fine-Tuning}
\label{sec:gift_finetuning}

\begin{figure*}[t]
	\centering
	\includegraphics[width=\textwidth]{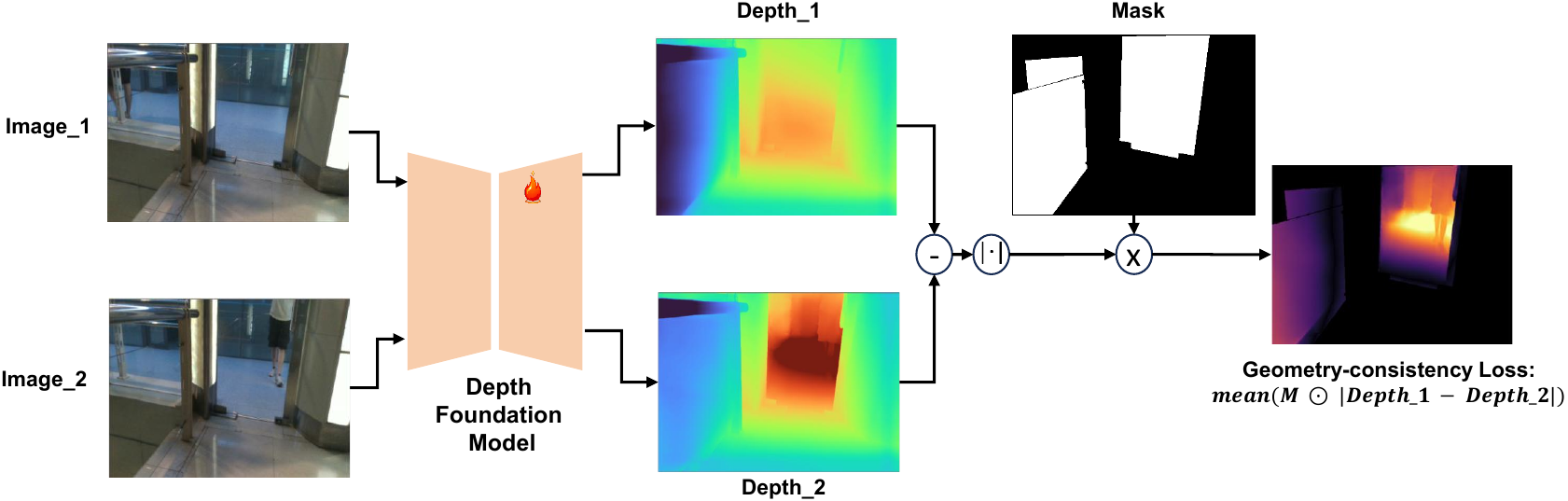}
	\caption{
		Pairwise illustration of geometry invariance. Observations with different
		non-Lambertian appearances should produce consistent target-region depth,
		measured by a normalized masked $L_1$ discrepancy.
	}
	\label{fig:pairwise_consistency}
\end{figure*}

\paragraph{Geometry-invariance principle.}
Observations from the same intervention group share the camera viewpoint and
physical target geometry but differ in reflected or transmitted appearance.
Their target-region depth predictions should therefore remain consistent.
As illustrated in Fig.~\ref{fig:pairwise_consistency}, given two observations
$x_i$ and $x_j$, let
$d_i=f_{\theta}(x_i)$ and $d_j=f_{\theta}(x_j)$ denote their predictions.
Using the shared target mask $M$ in this simplified pairwise illustration,
we define the geometry-consistency discrepancy as
\begin{equation}
	\ell_M(d_i,d_j)
	=
	\frac{
		\left\|M\odot(d_i-d_j)\right\|_1
	}{
		\|M\|_1
	},
	\label{eq:pairwise_consistency}
\end{equation}
where $\odot$ denotes element-wise multiplication. A lower value indicates
that the predicted target geometry is less sensitive to changes in
non-Lambertian appearance. This shared-mask formulation presents the core
idea; the complete-group objective below uses per-image masks and their
pairwise intersections.

\paragraph{Complete-group geometric consistency.}
GIFT extends the pairwise principle to all valid unordered pairs in an
intervention group. Given
$G=\{(x_i,m_i)\}_{i=1}^{N}$, let $d_i=f_{\theta}(x_i)$ and
$R_{ij}^{T}=m_i\cap m_j$ denote the prediction of observation $i$ and the
common target region of pair $(i,j)$, respectively. The mask intersection
reduces the influence of small registration and annotation-boundary errors.
We define the valid pair set as
\begin{equation}
	\mathcal U_T
	=
	\left\{
	(i,j)
	\,\middle|\,
	1\leq i<j\leq N,\;
	|R_{ij}^{T}|>0
	\right\},
	\label{eq:target_pair_set}
\end{equation}
and the complete-group consistency loss as
\begin{equation}
	\mathcal L_T
	=
	\frac{1}{|\mathcal U_T|}
	\sum_{(i,j)\in\mathcal U_T}
	\ell_{R_{ij}^{T}}(d_i,d_j),
	\label{eq:target_consistency}
\end{equation}
where $\ell$ is the normalized masked $L_1$ discrepancy defined in
Eq.~\ref{eq:pairwise_consistency}. Each prediction is computed once and
reused across all pairs involving that image.

\paragraph{Preventing collapse with frozen-model self-distillation.}
Consistency alone does not uniquely determine the correct physical depth:
any group-shared prediction, including a constant depth map, can minimize
$\mathcal L_T$ and cause the predicted geometry to collapse. To prevent
this degeneracy while preserving the pretrained depth prior, we constrain
the adapted prediction outside the annotated target using the frozen base
model. Let
\begin{equation}
	q_i
	=
	\operatorname{sg}
	\left[
	f_{\theta_0}(x_i)
	\right]
	\label{eq:frozen_base_prediction}
\end{equation}
denote the fixed base-model prediction, where $\operatorname{sg}[\cdot]$
stops gradients. These targets can be precomputed before fine-tuning.

The self-distillation loss is
\begin{equation}
	\mathcal L_{\mathrm{SD}}
	=
	\frac{2}{|\mathcal I_{\mathrm{SD}}|}
	\sum_{i\in\mathcal I_{\mathrm{SD}}}
	\ell_{\neg m_i}(d_i,q_i),
	\label{eq:self_distillation}
\end{equation}
where $\mathcal I_{\mathrm{SD}}$ contains images with valid non-target
pixels. The discrepancy follows Eq.~\ref{eq:pairwise_consistency} with the
region replaced by $\neg m_i$. Because it compares adapted and frozen
predictions for the same input, this term does not assume that non-target
regions are identical across different observations. It limits global
prediction drift and preserves general depth estimation outside the target.
The factor of two retains the relative contribution of the two endpoints
in the original pair-based formulation.

The complete GIFT objective is
\begin{equation}
	\mathcal L_{\mathrm{GIFT}}
	=
	\lambda_c\mathcal L_T
	+
	\lambda_d\mathcal L_{\mathrm{SD}},
	\label{eq:gift_objective}
\end{equation}
where $\lambda_c$ controls target-region adaptation and $\lambda_d$ controls
retention of the pretrained depth prior.

\paragraph{Efficient group-wise training.}
Algorithm~\ref{alg:group_training} summarizes the optimization procedure.
For each intervention group, GIFT forwards all observations once using
micro-batches and reuses the collected predictions to compute the
complete-group consistency loss and frozen-model self-distillation loss.
This avoids repeated forward passes for different image pairs and reduces
the model-forward cost from $N(N-1)/2$ pair evaluations to $N$ image
evaluations per group.

\begin{algorithm}[!t]
	\caption{Efficient GIFT training via group-wise optimization}
	\label{alg:group_training}
	\begin{algorithmic}[1]
		
		\Require Dataset $\mathcal D$, adapted model $f_{\theta}$,
		micro-batch size $K$
		
		\For{each intervention group $(X,M)$ with frozen targets $Q$}
		
		\State $D\gets\varnothing$
		
		\For{each micro-batch $X_S$ of size at most $K$ in $X$}
		
		\State $D_S\gets f_{\theta}(X_S)$
		
		\State $D\gets\operatorname{Concat}(D,D_S)$
		
		\EndFor
		
		\State Compute $\mathcal L_T$ from $D$ and $M$
		\Comment{Eq.~\ref{eq:target_consistency}}
		
		\State Compute $\mathcal L_{\mathrm{SD}}$ from $D$, $Q$, and $M$
		\Comment{Eq.~\ref{eq:self_distillation}}
		
		\State $\mathcal L_{\mathrm{GIFT}}\gets
		\lambda_c\mathcal L_T+
		\lambda_d\mathcal L_{\mathrm{SD}}$
		\Comment{Eq.~\ref{eq:gift_objective}}
		
		\State Update $\theta$ using $\nabla_{\theta}\mathcal L_{\mathrm{GIFT}}$
		
		\EndFor
		
	\end{algorithmic}
\end{algorithm}
\section{Experiments}
\label{sec:experiments}

\subsection{Experimental Setup}

\paragraph{Datasets.}

We evaluate GIFT on the validation split of GIFT-I and on the independent
Booster benchmark. The GIFT-I validation split contains 41 disjoint
intervention groups with 350 non-Lambertian input images. Each group
additionally provides one excluded surface-treated reference RGB image,
which is used only to construct a proxy depth reference and is never used
as input to the evaluated model.

We additionally evaluate on the balanced subset of the official Booster
training split~\citep{ramirez2024booster}. This subset contains 38 scenes
and 228 left-camera images. Among them, 188 images from 32 scenes contain
valid transparent-or-mirror (ToM) pixels. Booster provides stereo depth
annotations and is not used to train GIFT. We use \emph{ToM} to denote the
annotated transparent and mirror regions and \emph{All} to denote all valid
pixels in an image.We use \emph{Other} to denote valid pixels outside the annotated ToM region.

\paragraph{Metrics and affine-aligned evaluation.}
We evaluate geometry invariance using geometry-consistency error (GCE), which
measures the variation of predicted target geometry under controlled
appearance changes. For a validation set $\mathcal V$ of intervention groups,
we define
\begin{equation}
	\mathrm{GCE}
	=
	\frac{1}{|\mathcal V|}
	\sum_{g\in\mathcal V}
	\frac{1}{|\mathcal U_T^{(g)}|}
	\sum_{(i,j)\in\mathcal U_T^{(g)}}
	\ell_{R_{ij}^{T,(g)}}
	\left(
	d_i^{(g)},d_j^{(g)}
	\right),
	\label{eq:metric}
\end{equation}
where $\mathcal U_T^{(g)}$ is the valid unordered pair set of group $g$, and
$R_{ij}^{T,(g)}$ is the common target region of observations $i$ and $j$.
Lower GCE indicates stronger invariance to reflected or transmitted
appearance changes.

Because relative-depth predictions may differ in global scale and shift, we
affine-align each prediction to its reference using all valid pixels:
\begin{equation}
	\begin{aligned}
		(s^\star,t^\star)
		&=
		\arg\min_{s,t}
		\sum_{p\in\Omega_{\mathrm{valid}}}
		\left(s\hat y_p+t-y_p\right)^2, \\
		\tilde y_p
		&=
		s^\star\hat y_p+t^\star.
	\end{aligned}
	\label{eq:affine_alignment}
\end{equation}
The same transformation is applied to both the ToM and All regions. We then
report MAE, AbsRel, RMSE, and $\delta_1$, which evaluate aligned relative
geometry rather than direct metric-scale prediction.

For GIFT-I, $y$ is a proxy relative-depth label produced by the corresponding
frozen teacher model from the excluded surface-treated reference image. For
Booster, $y$ is the stereo-derived ground-truth depth used to evaluate the
monocular predictions. GIFT-I errors are reported in proxy units, whereas
Booster MAE and RMSE are reported in millimeters.

\paragraph{Implementation details.}
We adapt DAV2, VGGT, and Metric3D V2 using rank-8 LoRA while keeping the
majority of their pretrained parameters frozen. The corresponding depth
prediction head of each backbone is jointly optimized during fine-tuning.
All three backbones use the same input resolution and follow the group-wise
optimization procedure described in Algorithm~\ref{alg:group_training}.

All models are trained for 50 epochs using Adam with BF16 precision, a
learning rate of $2\times10^{-5}$, and no image augmentation. Each
intervention group produces one optimizer update, and validation is performed
after every epoch. The frozen version of each backbone is used to precompute
the self-distillation targets outside the annotated non-Lambertian regions.

\subsection{Quantitative Results}

\begin{table*}[t]
	\centering
	\small
	\setlength{\tabcolsep}{4.5pt}
	\begin{tabular}{crrrrrrrr}
		\toprule
		& \multicolumn{4}{c}{ToM}
		& \multicolumn{4}{c}{All} \\
		\cmidrule(lr){2-5}\cmidrule(lr){6-9}
		Method
		& MAE $\downarrow$
		& AbsRel $\downarrow$
		& RMSE $\downarrow$
		& $\delta_1$ $\uparrow$
		& MAE $\downarrow$
		& AbsRel $\downarrow$
		& RMSE $\downarrow$
		& $\delta_1$ $\uparrow$ \\
		\midrule
		VGGT-1B \citep{wang2025vggt}
		& 184.667 & 0.1476 & 227.309 & 84.43
		& 141.443 & 0.0779 & 230.101 & 93.67 \\
		DA V2-Large \citep{yang2024depthanythingv2}
		& 66.728 & 0.0737 & 81.810 & 93.21
		& 35.786 & 0.0307 & 58.634 & 98.25 \\
		Metric3D V2-Giant \citep{hu2024metric3dv2}
		& 234.962 & 0.1958 & 301.349 & 73.68
		& 167.029 & 0.1092 & 254.395 & 88.20 \\
		DA V3-Giant \citep{lin2026depthanything3}
		& 79.416 & 0.0717 & 99.924 & 94.77
		& 85.197 & 0.0523 & 140.604 & 97.12 \\
		MoGe-2-Large \citep{wang2025moge2}
		& 106.236 & 0.0946 & 121.600 & 93.39
		& 80.492 & 0.0515 & 133.929 & 97.44 \\
		\midrule
		\multirow{2}{*}{\textbf{VGGT-1B + GIFT}}
		& 62.465 & 0.0900 & 74.546 & 90.76
		& 30.803 & 0.0354 & 49.782 & 97.56 \\
		& \textbf{(-66.2\%)} & \textbf{(-39.1\%)}
		& \textbf{(-67.2\%)} & \textbf{(+7.5\%)}
		& \textbf{(-78.2\%)} & \textbf{(-54.6\%)}
		& \textbf{(-78.4\%)} & \textbf{(+4.2\%)} \\
		\addlinespace[2pt]
		\multirow{2}{*}{\textbf{DA V2-Large + GIFT}}
		& 50.542 & 0.0481 & 63.086 & 98.83
		& 38.897 & 0.0253 & 69.767 & 99.58 \\
		& \textbf{(-24.3\%)} & \textbf{(-34.7\%)}
		& \textbf{(-22.9\%)} & \textbf{(+6.0\%)}
		& (+8.7\%) & \textbf{(-17.6\%)}
		& (+19.0\%) & \textbf{(+1.4\%)} \\
		\addlinespace[2pt]
		\multirow{2}{*}{\textbf{Metric3D V2-Giant + GIFT}}
		& 217.195 & 0.1752 & 272.389 & 80.21
		& 141.615 & 0.0871 & 236.046 & 92.77 \\
		& \textbf{(-7.6\%)} & \textbf{(-10.5\%)}
		& \textbf{(-9.6\%)} & \textbf{(+8.9\%)}
		& \textbf{(-15.2\%)} & \textbf{(-20.3\%)}
		& \textbf{(-7.2\%)} & \textbf{(+5.2\%)} \\
		\bottomrule
	\end{tabular}
	\caption{
		Affine-aligned proxy-depth results on GIFT-I (backbone-specific proxy
		units). Bold method names denote GIFT variants, bold percentages mark
		improvements.
	}
	\label{tab:gifti_affine_comparison}
\end{table*}

\begin{table*}[t]
	\centering
	\small
	\setlength{\tabcolsep}{4.5pt}
	\begin{tabular}{crrrrrrrr}
		\toprule
		& \multicolumn{4}{c}{ToM}
		& \multicolumn{4}{c}{All} \\
		\cmidrule(lr){2-5}\cmidrule(lr){6-9}
		Method
		& MAE $\downarrow$
		& AbsRel $\downarrow$
		& RMSE $\downarrow$
		& $\delta_1$ $\uparrow$
		& MAE $\downarrow$
		& AbsRel $\downarrow$
		& RMSE $\downarrow$
		& $\delta_1$ $\uparrow$ \\
		& (mm) & & (mm) & (\%)
		& (mm) & & (mm) & (\%) \\
		\midrule
		VGGT-1B \citep{wang2025vggt}
		& 74.391 & 0.0740 & 94.121 & 94.55
		& 33.198 & 0.0287 & 52.115 & 98.78 \\
		DA V2-Large \citep{yang2024depthanythingv2}
		& 76.755 & 0.0798 & 100.060 & 95.29
		& 41.852 & 0.0373 & 58.771 & 99.37 \\
		Metric3D V2-Giant \citep{hu2024metric3dv2}
		& 70.922 & 0.0641 & 94.553 & 96.14
		& 43.241 & 0.0344 & 62.634 & 98.41 \\
		DA V3-Giant \citep{lin2026depthanything3}
		& 61.048 & 0.0527 & 76.492 & 96.62
		& 28.423 & 0.0234 & 43.906 & 99.40 \\
		MoGe-2-Large \citep{wang2025moge2}
		& 34.120 & 0.0387 & 44.410 & 99.36
		& 24.155 & 0.0211 & 35.383 & 99.71 \\
		\midrule
		\multirow{2}{*}{\textbf{VGGT-1B + GIFT}}
		& 61.585 & 0.0625 & 77.022 & 96.78
		& 29.057 & 0.0257 & 45.764 & 99.39 \\
		& \textbf{(-17.2\%)} & \textbf{(-15.6\%)}
		& \textbf{(-18.2\%)} & \textbf{(+2.4\%)}
		& \textbf{(-12.5\%)} & \textbf{(-10.7\%)}
		& \textbf{(-12.2\%)} & \textbf{(+0.6\%)} \\
		\addlinespace[2pt]
		\multirow{2}{*}{\textbf{DA V2-Large + GIFT}}
		& 51.450 & 0.0527 & 62.376 & 99.31
		& 41.591 & 0.0367 & 54.581 & 99.59 \\
		& \textbf{(-33.0\%)} & \textbf{(-33.9\%)}
		& \textbf{(-37.7\%)} & \textbf{(+4.2\%)}
		& \textbf{(-0.6\%)} & \textbf{(-1.5\%)}
		& \textbf{(-7.1\%)} & \textbf{(+0.2\%)} \\
		\addlinespace[2pt]
		\multirow{2}{*}{\textbf{Metric3D V2-Giant + GIFT}}
		& 70.376 & 0.0636 & 99.109 & 96.15
		& 36.194 & 0.0300 & 55.262 & 99.10 \\
		& \textbf{(-0.8\%)} & \textbf{(-0.7\%)}
		& (+4.8\%) & \textbf{(+0.01\%)}
		& \textbf{(-16.3\%)} & \textbf{(-12.7\%)}
		& \textbf{(-11.8\%)} & \textbf{(+0.7\%)} \\
		\bottomrule
	\end{tabular}
	\caption{
		Affine-aligned relative-depth results on Booster. MAE and RMSE are in
		mm and $\delta_1$ is in percent. Bold method names denote GIFT variants,
		and bold percentages mark improvements.
	}
	\label{tab:booster_affine_comparison}
\end{table*}

\paragraph{Results on GIFT-I.}
Table~\ref{tab:gifti_affine_comparison} shows that GIFT improves
non-Lambertian depth recovery for all three adapted backbones. ToM RMSE is
reduced by 67.2\% for VGGT-1B, 22.9\% for DA V2-Large, and 9.6\% for
Metric3D V2-Giant. VGGT-1B and Metric3D V2-Giant also improve across all
All-region metrics, while DA V2-Large exhibits mixed All-region errors,
indicating a stronger trade-off between target adaptation and retention.
Because the proxy references are backbone-specific, comparisons on GIFT-I
are made only between each backbone and its GIFT-adapted counterpart.

\paragraph{Results on Booster.}
Table~\ref{tab:booster_affine_comparison} reports results on the Booster
benchmark. GIFT reduces ToM RMSE by 18.2\% for VGGT-1B
and 37.7\% for DA V2-Large, while Metric3D V2-Giant shows mixed ToM results.
All-region RMSE decreases by 12.2\%, 7.1\%, and 11.8\% for VGGT-1B,
DA V2-Large, and Metric3D V2-Giant, respectively. Although MoGe-2-Large
achieves the strongest absolute results, the paired comparisons show
model-dependent gains rather than uniform improvement across all metrics.

\begin{figure*}[t]
	\centering
	\includegraphics[width=\textwidth]{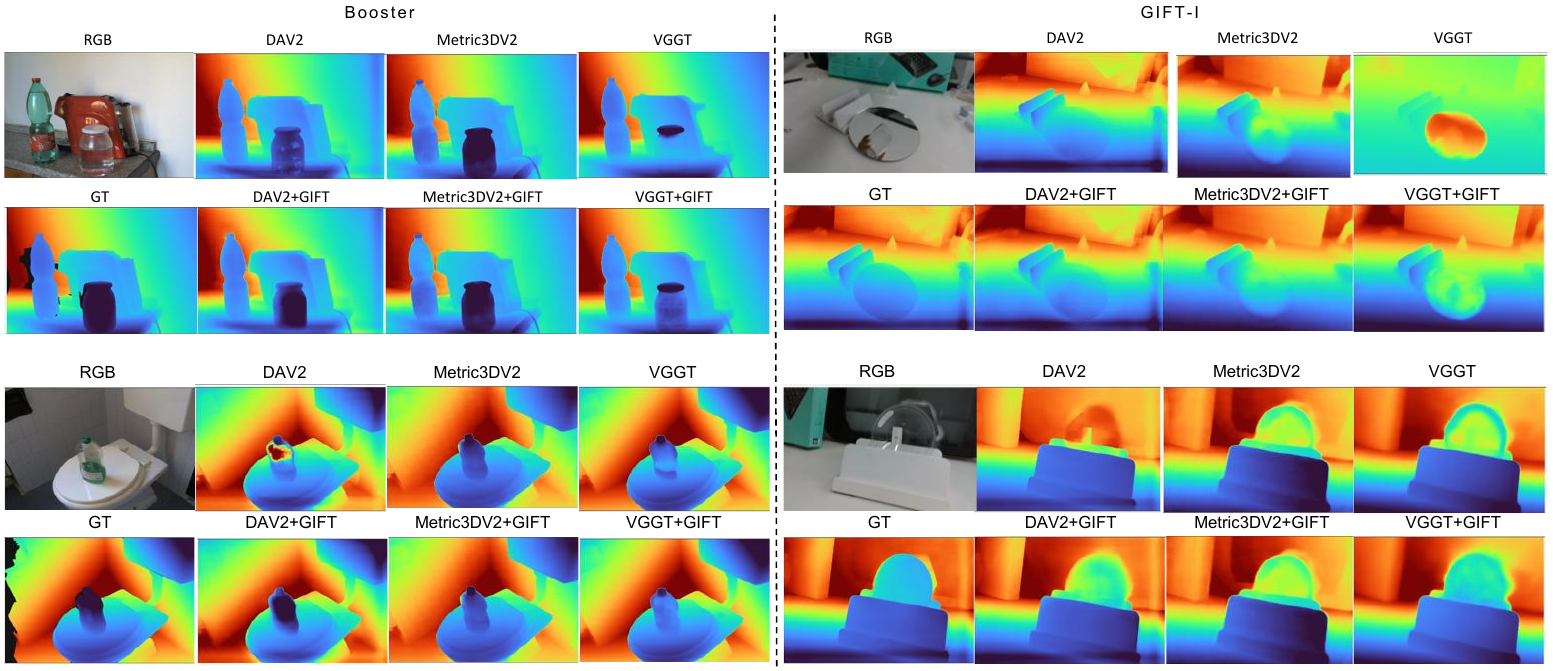}
	\caption{
		Qualitative comparison on Booster (left) and GIFT-I (right). For each
		example, the upper row shows RGB and base-model predictions, while the
		lower row shows the reference and corresponding GIFT-adapted
		predictions. GT denotes stereo ground truth on Booster and the
		surface-treated proxy reference on GIFT-I.
	}
	\label{fig:qualitative_comparison}
\end{figure*}

\paragraph{Qualitative comparison.}
Figure~\ref{fig:qualitative_comparison} presents selected examples from
GIFT-I and Booster. Relative to their corresponding base predictions, the
GIFT-adapted models recover target-region shapes and depth transitions that
more closely resemble the references. Additional zero-shot qualitative
comparisons and analysis are provided in the supplementary appendix.

\subsection{Effect of Geometry-Consistency Training}
\label{sec:effect_consistency}

\begin{figure}[t]
	\centering
	\includegraphics[width=0.7\columnwidth]{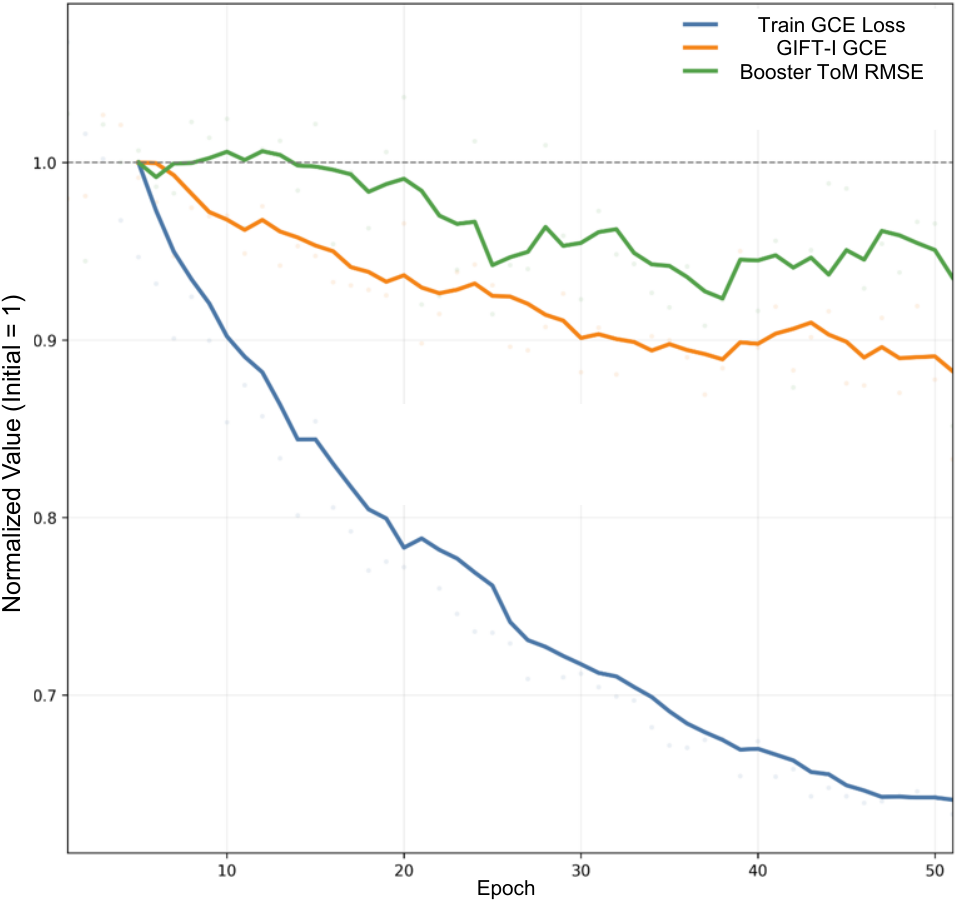}
	\caption{
		Smoothed, initial-normalized training GCE, GIFT-I GCE, and Booster ToM
		RMSE. Booster is evaluation-only.
	}
	\label{fig:effect_consistency}
\end{figure}

Figure~\ref{fig:effect_consistency} compares smoothed trajectories after
normalizing each metric by its initial value. Training GCE, GIFT-I validation
GCE, and Booster ToM RMSE all decrease overall. The aligned trends indicate
that the learned invariance generalizes to held-out groups and is accompanied
by improved target-region accuracy on an independent dataset. 

\subsection{Ablation of the Combined Loss Weights}
\label{sec:weight_ablation}

\begin{figure}[!t]
	\centering
	\includegraphics[width=\columnwidth]{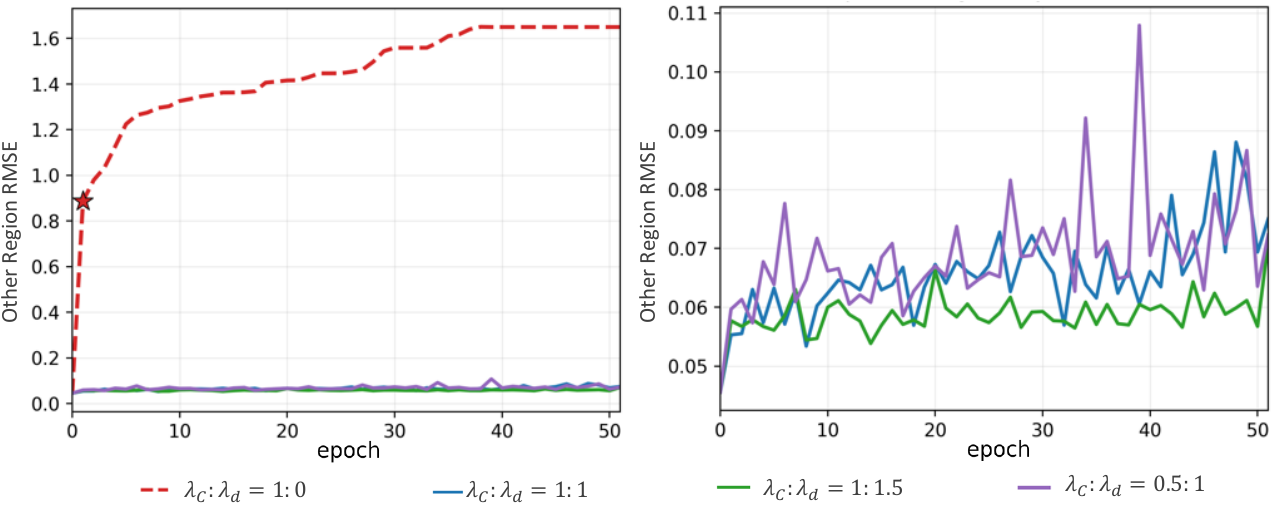}
	\caption{
		Other-region retention: removing self-distillation causes collapse
		(left), while weaker retention increases fluctuations (right).
	}
	\label{fig:loss_weight_dynamics}
\end{figure}

Figure~\ref{fig:loss_weight_dynamics} shows that $\lambda_d=0$ causes
Other-region RMSE to diverge because consistency alone admits degenerate
group-shared predictions. Among non-collapsed runs, weaker self-distillation
also produces larger fluctuations. Thus, self-distillation is necessary both
to prevent collapse and to preserve ordinary-region predictions.

\begin{table}[!t]
	\centering
	\small
	\setlength{\tabcolsep}{3pt}
	\begin{tabular}{ccccc}
		\toprule
		& \multicolumn{2}{c}{GIFT-I validation}
		& \multicolumn{2}{c}{Booster} \\
		\cmidrule(lr){2-3}\cmidrule(lr){4-5}
		$\lambda_c:\lambda_d$
		& GCE $\downarrow$
		& ToM RMSE $\downarrow$
		& ToM RMSE $\downarrow$
		& Other RMSE $\downarrow$ \\
		& & ($\times 10^{3}$)
		& (mm) & (mm) \\
		\midrule
		$1{:}1$
		& 0.0450 & \textbf{113.2} & 246.4 & 288.7 \\
		$1{:}1.5$
		& 0.0476 & 154.9 & \textbf{236.1} & \textbf{284.7} \\
		$1{:}2$
		& 0.0509 & 123.9 & 257.1 & 304.4 \\
		$2{:}1$
		& \textbf{0.0381} & 126.5 & 249.6 & 298.9 \\
		\bottomrule
	\end{tabular}
	\caption{
DAV2 loss-weight ablation at the minimum-GCE checkpoint. Booster is
evaluation-only; bold marks the best value.
	}
	\label{tab:weights}
\end{table}

Table~\ref{tab:weights} reports the minimum-GCE checkpoint from each
trajectory. The $2{:}1$ ratio gives the lowest GCE, while $1{:}1.5$ gives the
best Booster values; no setting dominates all criteria. We use $1{:}1$
because it achieves the lowest GIFT-I proxy ToM RMSE with competitive GCE,
without using Booster for selection. The collapsed $\lambda_d=0$ run is shown
only in Figure~\ref{fig:loss_weight_dynamics}.

\section{Limitations}

GIFT has two main limitations. First, although collecting grouped RGB images
is easier than acquiring reliable non-Lambertian depth, controlled capture,
registration, quality filtering, and mask annotation still require manual
effort, limiting the current scale and diversity of GIFT-I. Second, GIFT
suppresses appearance-dependent depth hallucinations without measured depth
supervision. Its accuracy therefore remains dependent on the pretrained
backbone, as reflected by the smaller gains obtained with Metric3D V2.
Future work will focus on automated data construction and combining geometry
invariance with additional depth supervision.

\section{Conclusion}

We presented GIFT, a geometry-invariant fine-tuning framework for
non-Lambertian monocular depth estimation. Using grouped observations from
GIFT-I, GIFT combines complete-group geometry-invariance supervision with
frozen-model self-distillation to reduce appearance-dependent depth
hallucinations while preserving the pretrained depth prior. We also introduced
GCE for depth-GT-free invariance evaluation. Experiments on GIFT-I and Booster
demonstrate improvements across Depth Anything V2, VGGT-1B, and Metric3D V2,
showing that controlled appearance intervention
provides a practical self-supervised signal for adapting depth foundation
models.



\clearpage
\setcounter{section}{0}
\setcounter{subsection}{0}
\setcounter{figure}{0}
\setcounter{table}{0}
\setcounter{equation}{0}
\setcounter{secnumdepth}{2}
\renewcommand{\thesection}{\Alph{section}}
\renewcommand{\thesubsection}{\thesection.\arabic{subsection}}
\renewcommand{\thefigure}{S\arabic{figure}}
\renewcommand{\thetable}{S\arabic{table}}
\renewcommand{\theequation}{S\arabic{equation}}
\def\GIFTARXIVCOMBINED{}
\ifdefined\GIFTARXIVCOMBINED\else
\documentclass[letterpaper]{article}

\usepackage[submission]{aaai2027}
\usepackage[hyphens]{url}
\usepackage{graphicx}
\urlstyle{rm}
\def\UrlFont{\rm}
\usepackage{natbib}
\usepackage{booktabs}
\usepackage{amsmath}
\usepackage{amssymb}
\usepackage{placeins}

\pdfinfo{
	/TemplateVersion (2027.1)
}

\setcounter{secnumdepth}{2}
\renewcommand{\thesection}{\Alph{section}}
\renewcommand{\thesubsection}{\thesection.\arabic{subsection}}

\renewcommand{\thefigure}{S\arabic{figure}}
\renewcommand{\thetable}{S\arabic{table}}
\renewcommand{\theequation}{S\arabic{equation}}

\begin{document}
\fi
	\raggedbottom
	
	\twocolumn[
	\begin{center}
		{\LARGE\bfseries
			GIFT: Geometry-Invariant Fine-Tuning for Non-Lambertian
			Monocular Depth Estimation\par}
		\vspace{0.6em}
		{\Large --- Supplementary Material ---\par}
	\end{center}
	\vspace{0.7em}
	]
	
	This supplementary material complements the main paper with further details
	on the construction and preprocessing of GIFT-I, the implementation of
	group-wise GIFT adaptation, and additional qualitative comparisons.
	Section A describes the capture equipment, appearance interventions, manual
	mask annotation, image registration, and dataset organization. Section B
	reports implementation and optimization details omitted from the main paper.
	Section C presents additional qualitative results on GIFT-I and Booster.
	
	\section*{Contents}
	
	\noindent\textbf{A. GIFT-I Dataset}\par
	\hspace{1.5em}\textbf{A.1} Data Collection and Annotation\par
	\hspace{1.5em}\textbf{A.2} Preprocessing and Dataset Overview\par
	
	\medskip
	\noindent\textbf{B. Training and Implementation Details}\par
	\hspace{1.5em}\textbf{B.1} Group-Wise Implementation\par
	\hspace{1.5em}\textbf{B.2} Model and Optimization Settings\par
	\hspace{1.5em}\textbf{B.3} Runtime and Efficiency\par
	
	\medskip
	\noindent\textbf{C. Additional Qualitative Comparisons}\par

	\section{GIFT-I Dataset}
	
	\subsection{Data Collection and Annotation}
	
	\paragraph{Capture setup.}
	Each GIFT-I group is captured using a fixed camera--target configuration.
	An Intel RealSense D435i camera is rigidly mounted on a tripod, and the
	non-Lambertian target remains stationary throughout the capture process.
	Only the RGB stream of the D435i is used.
	
	The D435i depth stream is deliberately excluded. Active depth measurements
	are often unreliable on reflective and transparent surfaces. More
	importantly, GIFT is designed to suppress appearance-dependent depth
	hallucinations under fixed physical geometry while preserving the geometric
	prior of a pretrained monocular model. It is not designed to eliminate the
	general model-to-real-world domain gap or to calibrate a monocular model to
	the metric output, noise characteristics, and biases of a particular RGB-D
	sensor. Using the D435i depth stream for training or evaluation would
	therefore conflate appearance-invariance adaptation with a separate
	sensor-domain adaptation problem.
	
\begin{figure}[t]
		\centering
		\includegraphics[width=0.8\columnwidth]{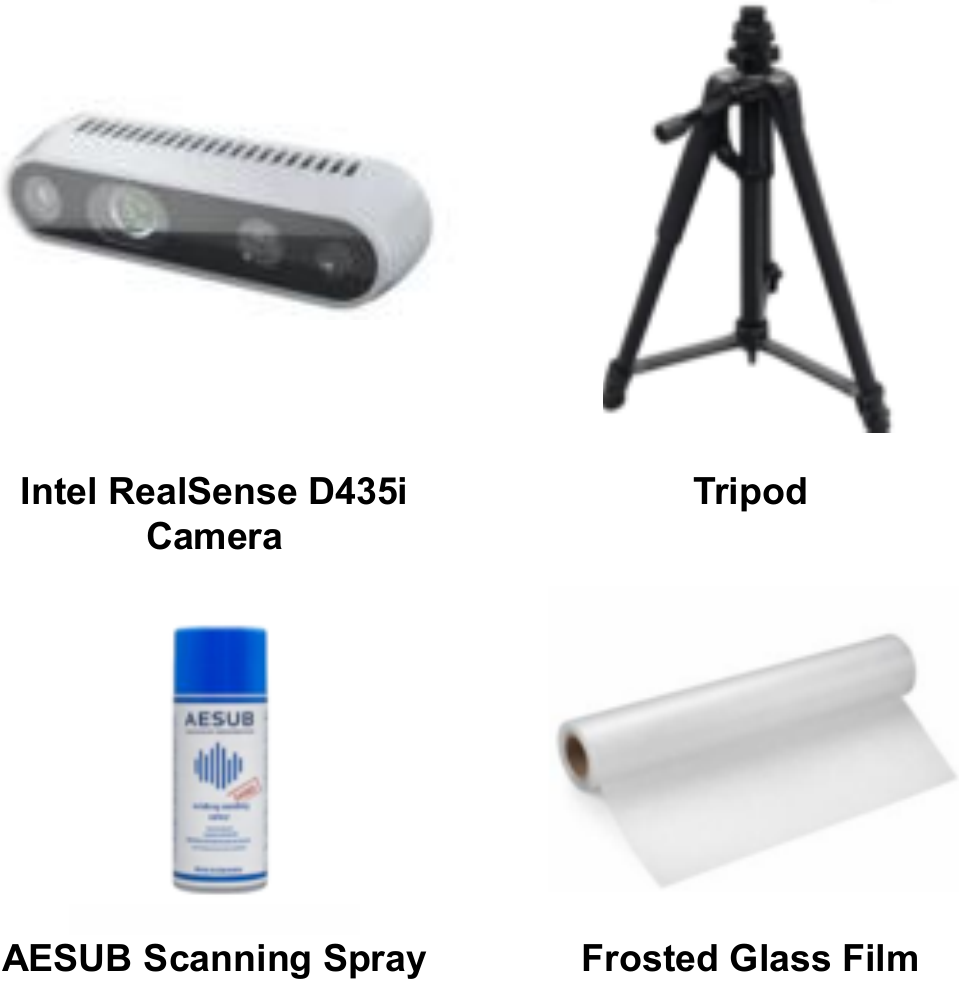}
		\caption{Equipment used for GIFT-I data collection and reference
			acquisition: an Intel RealSense D435i camera, a tripod, AESUB scanning
			spray, and frosted glass film.}
		\label{fig:capture_devices}
	\end{figure}

	Figure~\ref{fig:capture_devices} shows the equipment used for RGB collection
	and surface-treated reference acquisition, including the D435i camera, a
	tripod, AESUB scanning spray, and frosted glass film.
	
	\paragraph{Appearance interventions.}
	For transparent objects, the transmitted content behind the target is moved
	or replaced. For mirrors, the reflected scene content is changed. The camera
	pose and target geometry remain fixed, while the non-Lambertian appearance
	of the target changes with its reflected or transmitted environment. Other
	parts of the scene are kept approximately unchanged within the same group.
	
	These controlled interventions provide observations of the same physical
	surface geometry under different appearances, forming the geometry-invariance
	signal used by GIFT.
	
	\paragraph{Surface-treated reference RGB.}
	Each validation group additionally contains one surface-treated reference
	RGB image captured under the same camera pose and target geometry. The
	treatment directly changes the surface material and its image formation,
	replacing the original reflection- or transmission-dominated appearance with
	a more diffuse surface appearance.
	
	AESUB scanning spray and frosted glass film provide two practical treatment
	options. The choice primarily depends on the area and shape of the
	non-Lambertian surface. Scanning spray is suitable for relatively small or
	irregularly shaped targets. For large, approximately planar transparent
	surfaces, frosted glass film is preferred because it is more economical and
	material-efficient than coating the entire surface with scanning spray.
	
	The surface-treated reference RGB image is not included among the evaluated
	non-Lambertian inputs. It is used to construct the proxy reference for the
	corresponding validation group.
	
	\paragraph{Manual target-mask annotation.}
	The target region in every non-Lambertian RGB image is manually annotated
	with a binary mask. Masks are annotated independently for individual images
	rather than copied across a group, because small residual displacements may
	remain despite the fixed capture setup. Independent annotation also allows
	pairwise mask intersections to conservatively exclude uncertain boundary
	pixels after registration.

	\subsection{Preprocessing and Dataset Overview}
	
	\paragraph{Image registration.}
	All non-reference observations within each capture group are registered to
	the last naturally ordered non-reference RGB image. SIFT features are
	extracted using a contrast threshold of $0.08$ and an edge threshold of
	$10$. Candidate correspondences are first filtered using Lowe's ratio test
	with a threshold of $0.70$.
	
	A displacement-consistency filter then removes matches whose displacement
	differs from the median displacement by at least $20$ pixels. A homography
	is estimated using USAC-MAGSAC, with RANSAC used as a fallback. The
	reprojection threshold is $2$ pixels, the confidence is $0.9995$, and the
	maximum number of iterations is $5{,}000$.
	
	A registered frame is retained only when at least $90\%$ of the
	displacement-filtered correspondences are geometric inliers. The same
	homography is applied to the RGB image, target mask, and stored proxy depth.
	RGB images use linear interpolation, whereas masks and depth maps use
	nearest-neighbor interpolation.
	
	\paragraph{Mask processing and quality control.}
	All masks are binarized again after resizing and geometric transformation.
	For an image pair $(i,j)$, the common target region is defined as
	$m_i\cap m_j$. This intersection conservatively excludes pixels affected by
	residual registration errors or uncertain annotation boundaries.
	
	Dataset paths are stored relative to a configurable dataset root. During
	loading, images are grouped according to their capture directory, duplicate
	entries are removed, and group members are sorted using natural ordering.
	Every training RGB image must have a corresponding manually annotated mask
	and frozen proxy depth. Missing required files cause an explicit error
	rather than silent sample replacement. Groups containing fewer than two
	valid observations are rejected because they cannot provide an inter-image
	consistency target.
	
	\paragraph{Dataset composition.}
	GIFT-I contains 353 training groups with 3,268 RGB images. Group sizes range
	from 2 to 55, with a mean of 9.26 images. The validation set contains 41
	groups with 350 non-Lambertian RGB inputs. Each validation group additionally
	contains one surface-treated reference RGB image, resulting in 41 reference
	images.
	
	Table~\ref{tab:gift_i_statistics} summarizes the composition of GIFT-I.
	
	\begin{table}[t]
		\centering
		\small
		\setlength{\tabcolsep}{4.2pt}
		\begin{tabular}{lrrrr}
			\toprule
			Set & Groups & Inputs & References & Group size \\
			\midrule
			Training
			& 353 & 3,268 & 0 & 2--55 \\
			Validation
			& 41 & 350 & 41 & 3--23$^{\dagger}$ \\
			\bottomrule
		\end{tabular}
		\caption{Composition of GIFT-I.
			$^{\dagger}$The validation group size includes one surface-treated
			reference RGB image.}
		\label{tab:gift_i_statistics}
	\end{table}
	
	\paragraph{Frozen proxy generation.}
	For every training RGB image, the frozen version of the corresponding
	backbone predicts a proxy depth $q_i$. These proxies are generated once
	before adaptation and supervise only pixels outside the manually annotated
	target mask. No proxy depth is applied inside the non-Lambertian target
	region.
	
	For validation, the same frozen backbone predicts the depth of the
	surface-treated reference RGB image. The resulting prediction is used as
	the proxy reference for the non-Lambertian observations in the corresponding
	group. The GIFT-I reference is therefore model-derived rather than measured
	ground truth. Consequently, GIFT-I errors are used for before--after
	comparisons within the same backbone rather than for directly ranking
	different backbone families.
	
	All RGB images are resized to $476{\times}644$ using bicubic interpolation.
	Proxy inference is performed independently for each backbone in FP32 with a
	batch size of two. The resulting proxy predictions are retained in the native
	output scale of each backbone and are not interpreted as metric depth.
	GIFT-I evaluation follows the affine-alignment protocol defined in the main
	paper; the proxy values are therefore treated as backbone-specific relative
	depth.
	
\begin{figure*}[!t]
		\centering
		\includegraphics[width=\textwidth]{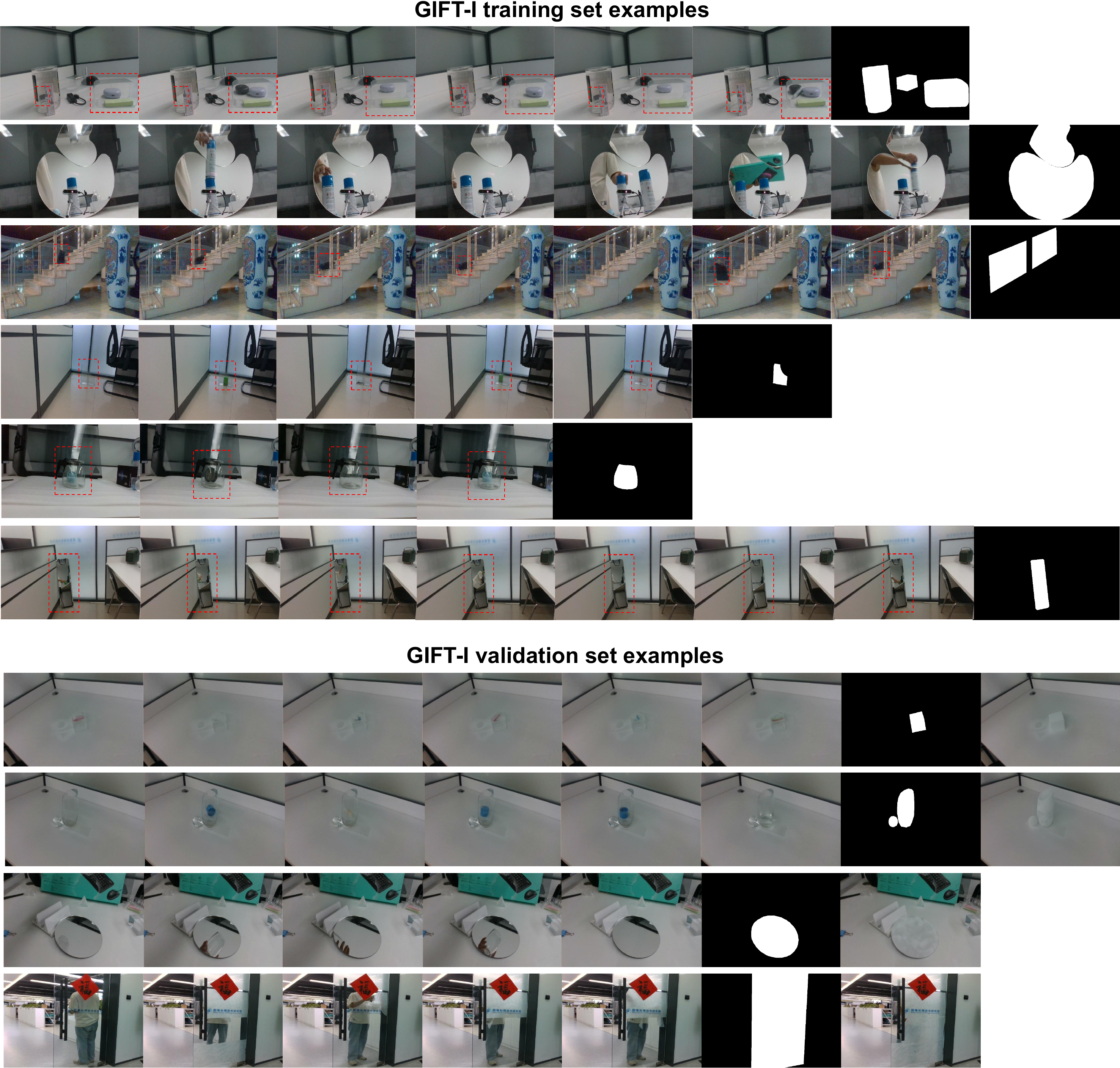}
		\caption{Representative complete groups from GIFT-I. Within each group,
			the camera pose and target geometry remain fixed while the reflected or
			transmitted content changes. Red boxes indicate the regions affected by
			the appearance intervention. All RGB observations belonging to each
			displayed group are shown. The surface-treated reference RGB image is
			used to construct the corresponding proxy reference.}
		\label{fig:supp_gift_groups}
	\end{figure*}

	Figure~\ref{fig:supp_gift_groups} presents representative complete groups
	from GIFT-I. For every displayed group, all captured RGB observations are
	shown. The camera pose and target geometry remain fixed within each group,
	while the reflected or transmitted content is deliberately changed. Red
	boxes identify the regions affected by the appearance intervention.
	
	\section{Training and Implementation Details}
	
	\subsection{Group-Wise Implementation}

	The pseudocode in the main paper presents the group-wise objective at a
	conceptual level. The implementation below realizes the same objective under
	bounded GPU memory using one snapshot forward and one replay forward per
	image.
	
	We use the complete-group geometry-consistency and frozen-model
	self-distillation objective defined in the main paper, with
	$\lambda_c:\lambda_d=1:1$. No additional loss is introduced.
	
	Each optimizer update consumes one complete intervention group. To limit
	memory usage, we implement the group-wise objective using
	snapshot-and-replay. Group predictions are first cached without gradient
	tracking in micro-batches of two.
	
	The images are then replayed with gradients enabled. For each micro-batch,
	its complete-group consistency contribution is computed against the cached
	group predictions, while self-distillation is evaluated against the
	precomputed frozen-model targets. Loss gradients are first computed with
	respect to the current depth predictions and then propagated through the
	current model computation graph.
	
	After each micro-batch backward pass, its computation graph is released,
	while parameter gradients are accumulated over the complete group. One
	optimizer update is applied after all group members have been processed.
	Thus, only detached group-level depth snapshots and accumulated parameter
	gradients are retained across micro-batches. If a group contains an odd
	number of images, the final single-image micro-batch is processed normally.
	
	Each epoch consists of one shuffled pass over all 353 training groups,
	resulting in 353 Adam updates. Frozen self-distillation targets are
	precomputed and loaded from disk during adaptation.

	\subsection{Model and Optimization Settings}
	
	We adapt DA V2-Large, VGGT-1B, and Metric3D V2-Giant using rank-8 LoRA
	modules. For each backbone, the corresponding depth prediction head is
	jointly optimized with the LoRA parameters, while the remaining pretrained
	parameters are frozen.
	
	All three models use the same optimization configuration. Input images are
	resized to $476{\times}644$, and each model is trained for 50 epochs using
	Adam with BF16 precision. The learning rate is fixed at
	$2{\times}10^{-5}$, with zero weight decay and no image augmentation. The
	geometry-consistency and self-distillation terms are equally weighted, with
	$\lambda_c:\lambda_d=1:1$. Each intervention group produces one optimizer
	update.
	
	\begin{table*}[!t]
		\centering
		\small
		\setlength{\tabcolsep}{7pt}
		\begin{tabular}{lrrrr}
			\toprule
			Implementation
			& Optimizer updates / epoch
			& Image forwards / epoch
			& Time / epoch
			& Relative speed \\
			\midrule
			Naive pair traversal
			& 20,339
			& 40,678
			& 122.4 min$^{\dagger}$
			& $1.0\times$ \\
			Snapshot-and-replay group-wise
			& 353
			& 6,536
			& 4.53 min
			& $27.0\times$ \\
			\bottomrule
		\end{tabular}
		\caption{Efficiency comparison between naive pair traversal and the
			snapshot-and-replay group-wise implementation for DA V2-Large on the
			353-group GIFT-I training set. The group-wise implementation uses one
			detached snapshot forward and one gradient-enabled replay forward per
			image, while reusing each snapshot across all associated pairwise
			comparisons. $^{\dagger}$The pair-traversal time is estimated from a
			measured average of $0.3612$ seconds per pair update.}
		\label{tab:efficiency_comparison}
	\end{table*}

	\subsection{Runtime and Efficiency}
	
	\paragraph{Runtime environment.}
	Training time is measured on one NVIDIA GeForce RTX 4090 with 24~GB of GPU
	memory. The software environment uses Python 3.10.6, PyTorch 2.5.1 with
	CUDA 12.1, cuDNN 9.1, and xFormers 0.0.28.post3. The reported runtime covers
	only the 353 complete-group training updates in one epoch; evaluation and
	model I/O are excluded.
	
	Under the snapshot-and-replay group-wise implementation, one epoch takes
	4.53 minutes for DA V2-Large, 5.90 minutes for VGGT-1B, and 6.94 minutes for
	Metric3D V2-Giant.
	
	\paragraph{Pair traversal versus group-wise optimization.}
	The 353 training groups contain 20,339 unordered image pairs. A naive
	pair-traversal implementation processes every pair independently, requiring
	20,339 optimizer updates and 40,678 image forward passes per epoch.
	
	The snapshot-and-replay group-wise implementation performs one no-gradient
	snapshot forward and one gradient-enabled replay for each of the 3,268
	training images. It therefore requires 6,536 image forward passes and 353
	optimizer updates per epoch. The detached snapshots are reused across all
	pairwise comparisons involving the corresponding image.
	
	Table~\ref{tab:efficiency_comparison} summarizes the comparison. Relative to
	naive pair traversal, the group-wise implementation reduces the number of
	optimizer updates by $57.6\times$ and the number of image forward passes by
	$6.22\times$. The reported $27.0\times$ wall-clock speedup is an engineering
	estimate because the pair-traversal runtime is extrapolated from its measured
	per-update throughput. The comparison includes both prediction reuse and the
	reduced number of optimizer updates.

	\section{Additional Qualitative Comparisons}
	
\begin{figure*}[t]
		\centering
		\includegraphics[width=\textwidth]{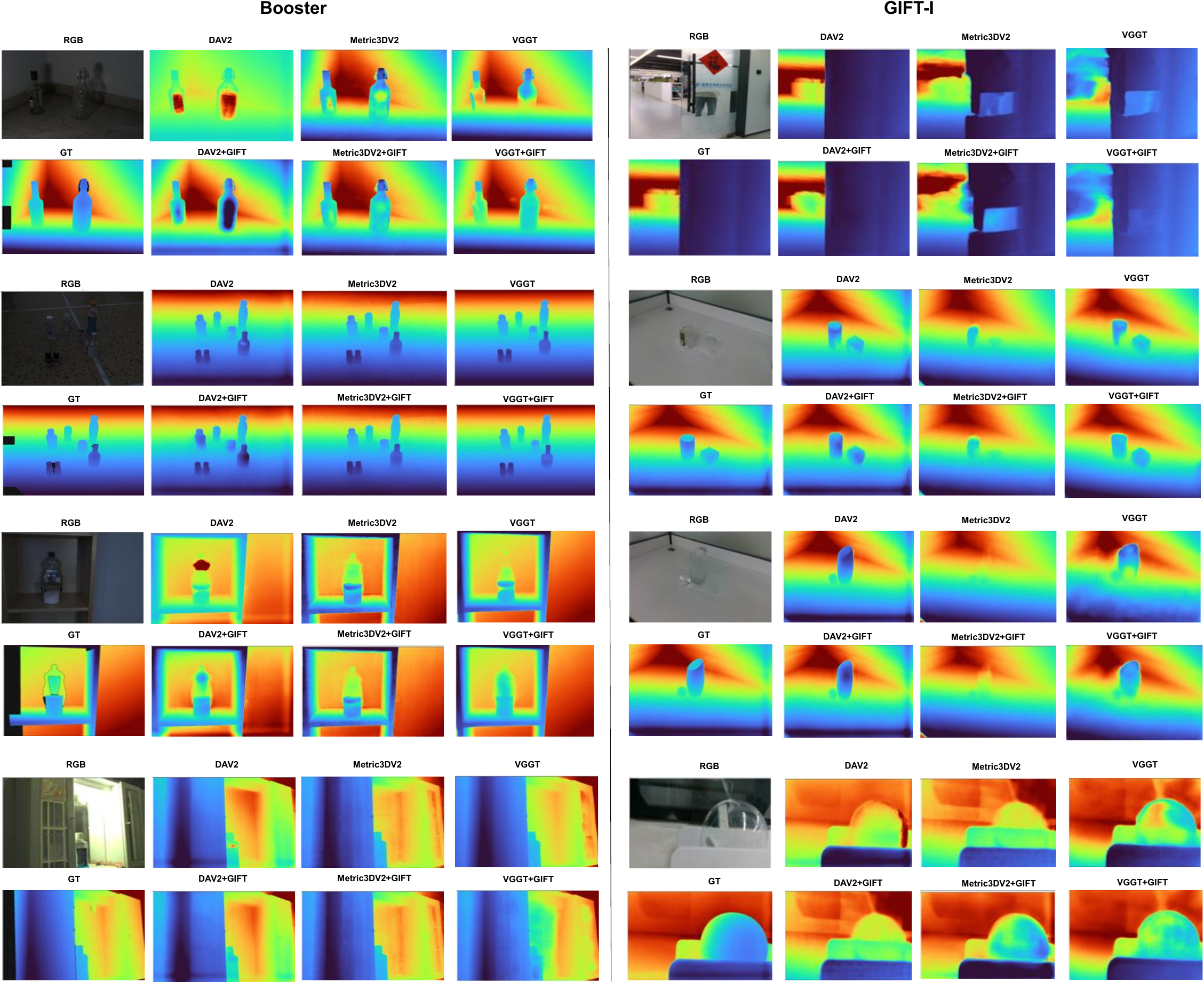}
		\caption{Additional qualitative comparisons on GIFT-I and Booster.
			For compact presentation, the reference depth is labeled ``GT'' in both
			datasets. On Booster, GT denotes stereo-derived ground-truth depth. On
			GIFT-I, GT denotes the corresponding backbone-specific proxy reference
			generated from the surface-treated reference RGB image and is not
			measured ground truth. For each sample, the base-model and GIFT-adapted
			predictions are affine-aligned to the corresponding reference and
			visualized using the same depth range. Invalid stereo pixels are shown
			only in the Booster GT image and are not transferred to the monocular
			predictions.}
		\label{fig:supp_qualitative_comparisons}
	\end{figure*}

	Figure~\ref{fig:supp_qualitative_comparisons} presents additional qualitative
	comparisons on GIFT-I and Booster. The visualization follows the affine
	alignment protocol defined in the main paper. For each sample, the base-model
	and GIFT-adapted predictions are aligned to the corresponding reference and
	visualized using the same depth range.
	
	For compact presentation, the reference-depth row is labeled ``GT'' for both
	datasets in the figure. Its meaning is dataset-dependent. On Booster, GT
	denotes stereo-derived ground-truth depth. On GIFT-I, GT denotes the
	corresponding backbone-specific proxy reference generated from the
	surface-treated reference RGB image; it is not measured ground truth.
	
	Invalid stereo pixels on Booster are displayed only in the GT image and are
	excluded from evaluation. They are not copied into, removed from, or
	overlaid on the dense monocular predictions.
	
	\subsection{Zero-Shot Comparison}
	
\begin{figure*}[t]
		\centering
		\includegraphics[width=\textwidth]{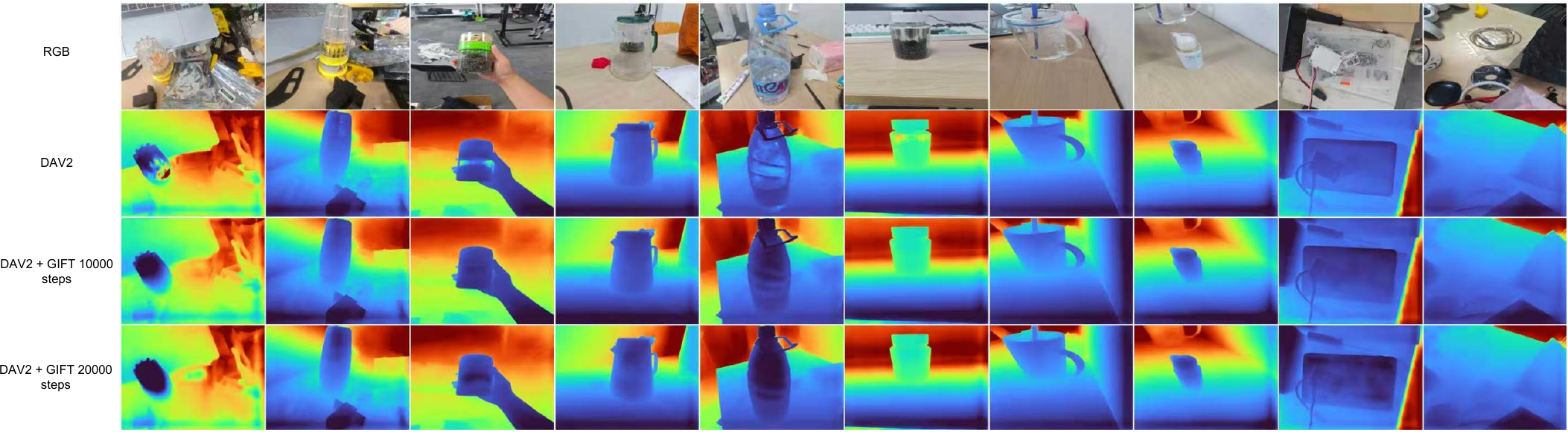}
		\caption{Zero-shot comparison of DA V2-Large before adaptation and after
			10,000 and 20,000 GIFT optimization steps. With longer adaptation,
			the predictions become progressively more conservative:
			appearance-dependent depth structures are increasingly suppressed,
			resulting in smoother and less appearance-sensitive estimates of the
			non-Lambertian surfaces. These examples are used only for qualitative
			analysis and are not included in GIFT-I or Booster.}
		\label{fig:supp_zero_shot}
	\end{figure*}

	Figure~\ref{fig:supp_zero_shot} compares the frozen DA V2-Large model with
	models adapted for 10,000 and 20,000 optimization steps on a set of zero-shot
	images that are not included in GIFT-I or Booster. These images are used only
	for qualitative visualization and do not participate in training or
	quantitative evaluation. The 20,000-step checkpoint is obtained from an
	extended diagnostic run beyond the 50-epoch schedule used for the quantitative
	experiments. Neither diagnostic checkpoint is used in the reported
	quantitative results.
	
	As the adaptation duration increases, the predicted geometry becomes
	progressively more conservative. Compared with the frozen model, the
	10,000-step model reduces appearance-dependent depth structures in the
	non-Lambertian regions, while the 20,000-step model further suppresses such
	variations and produces smoother, less appearance-sensitive surface
	predictions. This trend suggests that longer geometry-invariance adaptation
	increases the model's preference for stable surface geometry over depth
	details inferred from reflected or transmitted content.
	
	However, a more conservative prediction is not necessarily uniformly better.
	Excessive adaptation may also suppress plausible geometric details.
	Therefore, the comparison illustrates the effect of adaptation duration
	rather than establishing that more optimization steps always improve depth
	accuracy.
	
\ifdefined\GIFTARXIVCOMBINED\else
\end{document}
\fi

\end{document}